\documentclass{article}

\usepackage[preprint]{neurips_2023}

\usepackage[utf8]{inputenc} % allow utf-8 input
\usepackage[T1]{fontenc}    % use 8-bit T1 fonts
\usepackage{url}            % simple URL typesetting
\usepackage{booktabs}       % professional-quality tables
\usepackage{amsfonts,amsmath,amssymb}       % blackboard math symbols
\usepackage{nicefrac}       % compact symbols for 1/2, etc.
\usepackage{microtype}      % microtypography
\usepackage{xcolor}         % colors
\usepackage{graphicx}
\usepackage{comment}
\usepackage{wrapfig}
\usepackage{multirow}

\usepackage[pagebackref,breaklinks,colorlinks,bookmarks=false]{hyperref}

\makeatletter\renewcommand\paragraph{\@startsection{paragraph}{4}{\z@}
  {.5em \@plus1ex \@minus.2ex}{-.5em}{\normalfont\normalsize\bfseries}}\makeatother

\newcommand{\name}{ProRes}

\usepackage{authblk}

\title{ProRes: Exploring Degradation-aware Visual Prompt for Universal Image Restoration}

\newcommand{\etal}{\textit{et al}.}
\newcommand{\ie}{\textit{i}.\textit{e}.}
\newcommand{\eg}{\textit{e}.\textit{g}.}

\author[ ]{\textbf{Jiaqi Ma}$^{1}$\thanks{Equal contribution. This work is done when Jiaqi Ma was an intern at Horizon Robotics. Corresponding to Lefei Zhang <zhanglefei@whu.edu.cn>.}}
\author[ ]{\textbf{Tianheng Cheng}$^{2*}$}
\author[ ]{\textbf{Guoli Wang}$^{3}$}
\author[ ]{\textbf{Qian Zhang}$^{3}$}
\author[ ]{\textbf{Xinggang Wang}$^{2}$}
\author[ ]{\textbf{Lefei Zhang}$^{1}$}

\affil[1]{School of Computer Science, Wuhan University}
\affil[2]{School of EIC, Huazhong University of Science \& Technology}
\affil[3]{Horizon Robotics}
\affil[ ]{\tt\small \{jiaqima,zhanglefei\}@whu.edu.cn}
\affil[ ]{\tt\small \{thch,xgwang\}@hust.edu.cn}
\affil[ ]{\tt\small \{guoli.wang,qian01.zhang\}@horizon.ai}

\begin{document}

\maketitle
\begin{figure}[h]
  \centering
  \includegraphics[width=\linewidth]{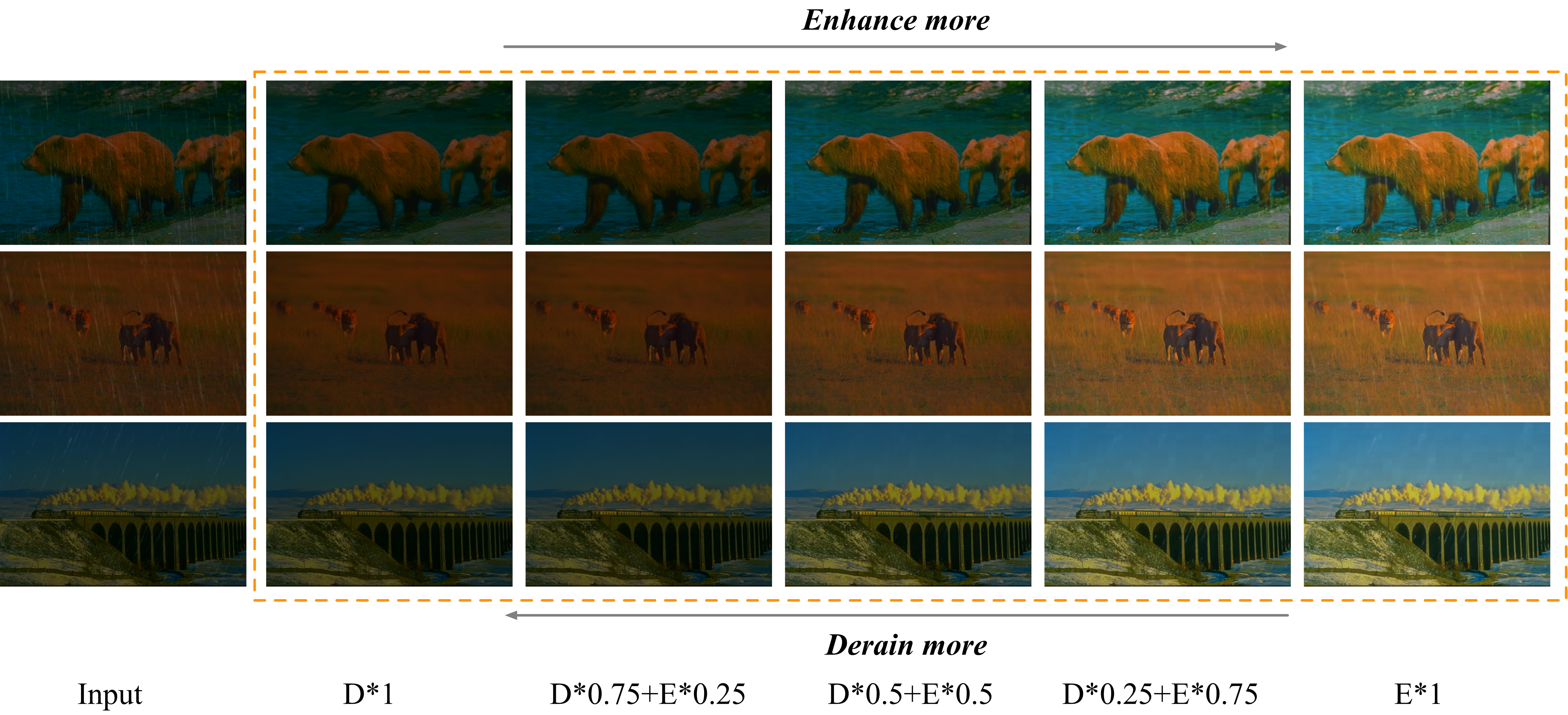}
  \vspace{-15pt}
  \caption{We show the visualization results processed by \textbf{ProRes} from images of mixed types of degradation, \ie, low-light and rainy.
  \name{} adopts two visual prompts for low-light enhancement (E) and deraining (D) and combines the two visual prompts by linear weighted sum, \ie, $\alpha \textrm{D} + (1-\alpha) \textrm{E}$, to control the restoration process.}
  \label{fig:S3_combine}
\end{figure}
\begin{abstract}

Image restoration aims to reconstruct degraded images, \eg, denoising or deblurring. Existing works focus on designing task-specific methods and there are inadequate attempts at universal methods.
However, simply unifying multiple tasks into one universal architecture suffers from uncontrollable and undesired predictions.
To address those issues, we explore prompt learning in universal architectures for image restoration tasks.
In this paper, we present \textbf{Degradation-aware Visual Prompts}, which encode various types of image degradation, \eg, noise and blur, into unified visual prompts.
These degradation-aware prompts provide control over image processing and allow weighted combinations for customized image restoration as shown in Fig.~\ref{fig:S3_combine}.
We then leverage degradation-aware visual \textbf{Pro}mpts to establish a controllable and universal model for image \textbf{Res}toration, called \textbf{ProRes}, which is applicable to an extensive range of image restoration tasks.
\name{} leverages the vanilla Vision Transformer (ViT) without any task-specific designs. 
Furthermore, the pre-trained \name{} can easily adapt to new tasks through efficient prompt tuning with only a few images.
Without bells and whistles, \name{} achieves competitive performance compared to task-specific methods and experiments can demonstrate its ability for controllable restoration and adaptation for new tasks.
The code and models will be released in \url{https://github.com/leonmakise/ProRes}.
\end{abstract}

\section{Introduction}

Image restoration, as the fundamental challenge in the field of computer vision, aims to reconstruct high-quality images from degraded images which suffer from noise, blur, compression artifacts, and other distortions.
Those kinds of low-level vision tasks are critical in a wide range of applications, including general vision perception, medical imaging, and satellite imaging. 

\begin{figure}
    \centering
    \includegraphics[width=\linewidth]{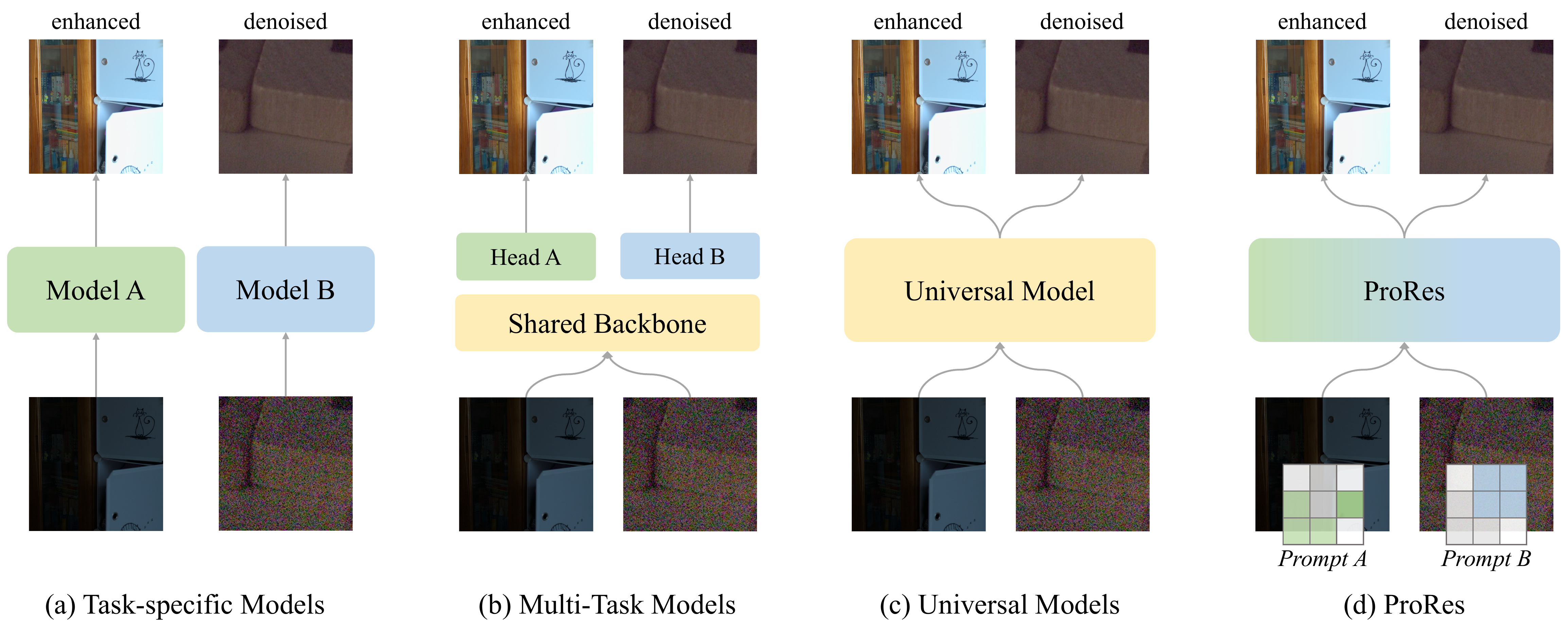}
    \vspace{-10pt}
    \caption{\textbf{Conceptual comparison with previous approaches.} (a) Task-specific models design specialized architectures and strategies for different tasks, \eg, Model-A for low-light enhancement and Model-B for image denoising. (b) Multi-task models adopt a shared backbone for image feature extraction and leverage multiple task-specific heads for different tasks. (c) Universal models adopt mixed inputs without any task-specific indicator. (d) The proposed \name{} adopts input images with degradation-aware visual prompts for specific targets.}
    \label{fig:paradigm_comparison}
\end{figure}

In recent years, deep learning methods~\cite{DBLP:conf/eccv/ChenCZS22, DBLP:journals/pami/ZamirAKHKYS23, DBLP:conf/wacv/MehriAS21, DBLP:conf/cvpr/WangCBZLL22} have revolutionized the field of image restoration, achieving high accuracy rates and improving the quality of restored images. 
Prevalent works concentrate on carefully designing task-specific methods and have shown promising results in various low-level image restoration tasks, \eg, denoising, deraining, and deblurring. 
However, the task-specific methods have limited transfer ability on new datasets and still require specific designs for adapting to other tasks, as shown in Fig.~\ref{fig:paradigm_comparison} (a).
Moreover, it's much more challenging for task-specific methods to process images of different corruptions.
Developing general approaches for simultaneously processing different corruptions is necessary for nowadays applications.
% transfer ability
% can not process various tasks sim

Recently, multi-task learning~\cite{DBLP:conf/cvpr/LiLHW0022, DBLP:conf/cvpr/ChenHTYDK22, DBLP:journals/pami/YuWDTL22} has been explored for processing images with different types of degradation by sharing the backbone and designing task-specific heads, as shown in Fig.~\ref{fig:paradigm_comparison} (b).
Despite its success in image restoration, multi-task methods with shared parameters still suffer from the task-interference problem~\cite{UniperceiverMOE} that different tasks might have different optimization directions. 

Inspired by the pioneering works on universal approaches for image recognition~\cite{Uni-Perceiver,UniperceiverMOE} and image segmentation~\cite{mask2former,Jain2022OneFormerOT}, we aim to investigate the universal architectures for image restoration.
In this paper, we formulate different image restoration tasks into a universal architecture, as illustrated in Fig.~\ref{fig:paradigm_comparison} (c), in which images from mixed tasks are fed into the universal model for unified training.
Nonetheless, the inference using the universal model is uncontrollable or unpredictable, as it cannot guarantee that the output will meet our expectations without any task-specific indicator.

To mitigate the above issues, we present Degradation-aware Visual Prompts as parametric identifiers for different types of degradation, \eg, a visual prompt for "low-light enhancement".
Specifically, we define a series of visual prompts with the same size as the images.
And then we add the selected visual prompt to the degraded image as a prompted image and feed it into the universal architecture for the desired restored image, as shown in Fig.~\ref{fig:paradigm_comparison} (d).
Further, we incorporate the degradation-aware prompts into a simple universal architecture, \ie, a vanilla Vision Transformer~\cite{DBLP:conf/iclr/DosovitskiyB0WZ21} with a pixel decoder, for universal image restoration, and present a novel and versatile framework named \textbf{\name{}}.
Compared to previous task-specific or multi-task approaches, \name{} with task-agnostic designs can be trained with various tasks and can easily process various types of degradation.
% \name{} is pre-trained on the joint dataset with various tasks.
% In this paper, we aim to investigate what prompt-based universal \name{} can do for image restoration.
% In this paper, we aim to show two superior 
In this paper, we aim to demonstrate two superior capabilities of the proposed \name{}: (1) \name{} has strong control ability, allowing us to combine different prompts with different weights according to our needs and obtain the desired output; (2) \name{} possesses exceptional transferability, where we can quickly and cost-effectively adapt \name{} to new tasks or datasets using simple prompt tuning.

To verify the effectiveness of the proposed \name{}, we construct a joint image restoration dataset of several tasks including denoising, low-light enhancement, deraining, and deblurring, and train \name{} on it.
% Then we directly evaluate \name{} on different tasks without extra fine-tuning.
Without bells and whistles, \name{}, as a universal approach, achieves competitive performance on various benchmarks, \eg, SSID~\cite{abdelhamed2018high} and LOL~\cite{wei2018deep}, compared to well-established and carefully-designed methods.
Surprisingly, qualitative and quantitative results can demonstrate the control ability of the proposed degradation-aware visual prompts, which can be combined with weights to generate desired restored images.
Moreover, experimental results on haze removal show that the proposed \name{} with prompt tuning can be efficiently and effectively adapted to new tasks.
And we believe large-scale datasets can further boost the capability of the proposed \name{} for both controllable image restoration and new-task adaption.
Our contribution can be summarized as follows:
\begin{itemize}
    \item We present degradation-aware visual prompts for universal image restoration which provide control over image processing given any degraded image.
    \item We propose \name{} to address universal image restoration with degradation-aware prompts, which is the first prompt-based versatile framework for image restoration.
    \item We additionally propose the effective and efficient prompt tuning with \name{} to adapt for new tasks or new datasets without fine-tuning \name{}.
    \item The proposed \name{} obtains competitive results compared to task-specific methods on various benchmarks. We hope the simple  yet effective \name{} can serve as a solid baseline for universal image restoration and facilitate future research.
\end{itemize}

\section{Related Works}

\subsection{Multi-Task Learning for Image Restoration}
Multi-task learning (MTL) has emerged as a promising paradigm by leveraging shared information across multiple related tasks. For image restoration tasks, MTL is employed to solve multiple related tasks simultaneously, such as denoising and super-resolution. The shared structure or feature representation across these tasks can often lead to enhanced performance compared to models trained on individual tasks. One notable work is the AIRNet \cite{DBLP:conf/cvpr/LiLHW0022}, which incorporated MTL by feeding several corruption. Besides, \cite{DBLP:conf/cvpr/ChenHTYDK22} constrains relationships of tasks by the multi-contrastive regularization. Path-Restore \cite{DBLP:journals/pami/YuWDTL22} offers a multi-path CNN to select an appropriate route for each image region.

However, multi-task learning for image restoration is achieving a balance between learning shared features and preserving task-specific features. The common framework contains several encoders for task selection or decoders for variable outputs. Our work addresses this challenge by integrating degradation-aware visual prompts, enabling us to control the task-specific aspects while still benefiting from the shared features.

\subsection{Universal Foundation Models}
Foundation models aim to serve as a shared basis for multiple tasks. The motivation behind such universal models is to leverage the commonalities among various tasks and modalities to improve efficiency and performance. Notable instances include the Vision Transformer (ViT) \cite{DBLP:conf/iclr/DosovitskiyB0WZ21} and BERT \cite{DBLP:conf/naacl/DevlinCLT19}, which have been used across various vision or language tasks. Perceiver \cite{DBLP:conf/icml/JaegleGBVZC21} and Perceiver-IO \cite{DBLP:conf/iclr/JaegleBADIDKZBS22} provide universal solutions to natural language and visual understanding processing. Then Uni-Perceiver \cite{DBLP:conf/cvpr/ZhuZLWLWD22} is extended to generic perception tasks and Uni-Perceiver-MoE \cite{DBLP:conf/nips/ZhuZWWLWD22} discusses the main factor to performance degradation. Besides, for a series of similar tasks, there are some universal models based on various of network architectures \cite{DBLP:journals/corr/abs-2304-08870, DBLP:journals/tmm/SongLLNLL21, DBLP:journals/corr/abs-2211-02043, DBLP:conf/iccv/MaHWQG21}. 

Those works also reveal the potential of the universal foundation model. Universal foundation models for low-level vision tasks, which can handle tasks such as image denoising, deraining, and low-light enhancement with a single model architecture, are still lacking in exploration. \cite{DBLP:conf/cvpr/ChenHTYDK22, DBLP:conf/nips/BarGDGE22, painter} involve in inpainting and denoising. However, there still exists no universal model, especially for low-level vision tasks, which can provide controllable predictions.

\subsection{Visual Prompt Learning}
Inspired by the success of prompt learning in natural language processing, recent works attempt to adapt prompt learning to several visual tasks. MAE-VQGAN \cite{DBLP:conf/nips/BarGDGE22} gives the grid-like input during the inference phase, and automatically inpaints blank areas by a pre-trained generative model. Then a supervised version of visual prompt learning named Painter \cite{painter} removes generative models like VQGAN, and calculates the regression loss by tokens from the vision transformer. These approaches usually involve providing the model with additional input or cues to guide the model's reasoning and attention. Some studies have demonstrated the effectiveness of visual prompts in tasks like image classification \cite{radford2021learning}, object detection \cite{cao2020open}, and visual question answering \cite{li2021align}.

\section{Methodology}
In this section, we begin by introducing degradation-aware visual prompts designed for various image restoration tasks. Then, we present our approach called \name{}, which integrates degradation-aware prompts to achieve universal image restoration. Furthermore, we demonstrate the versatility of \name{} by adapting it to new tasks through prompt tuning.

\begin{figure}
    \centering
    \includegraphics[width=1.0\linewidth]{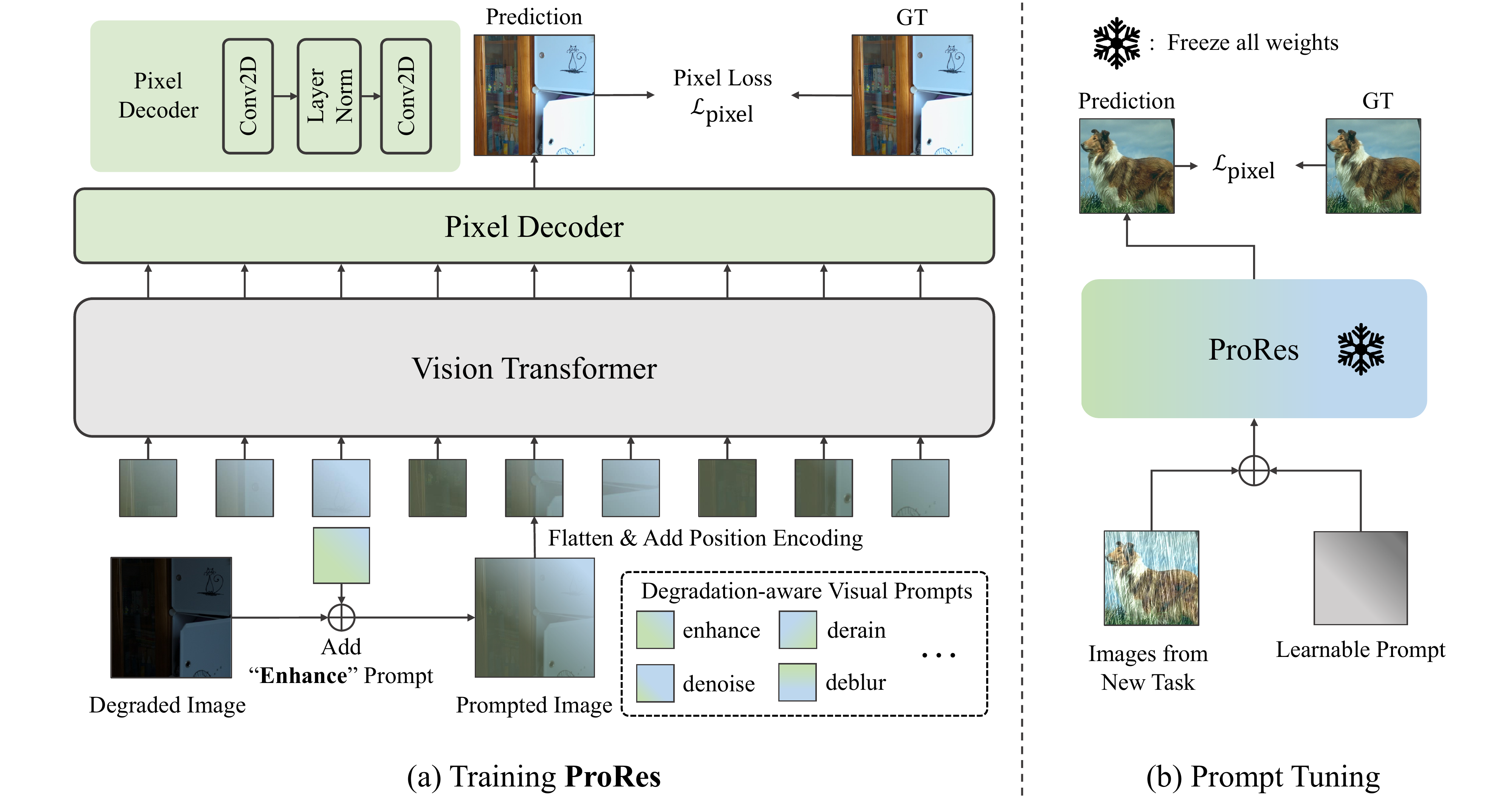}
    \vspace{-10pt}
    \caption{\textbf{Overall Pipeline of \name{}.} \textbf{(a) Training \name{}}: we add the target visual prompt to the input image and flatten the prompted image into patches. We leverage a vision transformer, \ie, ViT-Large, as the image encoder and adopt a simple pixel decoder to generate the restored image. Then we adopt pixel loss to optimize \name{}. \textbf{(b) Prompt Tuning}: we freeze the weights of \name{} and randomly initialize the learnable prompts for new tasks or new datasets.}
    \vspace{-20pt}
    \label{fig:main}
\end{figure}

\subsection{Degradation-aware Visual Prompts}

\paragraph{Prompt Design.}
% As we adopt ViT-Large as the backbone of \name{}, it is natural to use token-based prompts. 
Previous methods~\cite{painter, DBLP:conf/nips/BarGDGE22} utilize grid-like combinations of inputs and treat tasks as inpainting problems. However, the grid-like layout increases computational costs since the input becomes four times larger than the original image. Additionally, when using different predefined sets of prompts, the outputs may significantly vary for a pre-trained model.

Nevertheless, we concur~\cite{painter} that the format of visual prompts should resemble images, even if they do not possess visual meaning. In line with this, we define each visual prompt with a shape of $H \times W \times 3$, which matches the shape of the input images. The image-like format can be seamlessly integrated with any natural image and provides ample information as a prompt. 

Considering the limitations of the grid-like layout, the visual prompts are added directly to input images, which allows it to overcome the limitations of grid-like combinations and achieve more efficient and consistent results. For each task, a single visual prompt is utilized to capture and represent the task-specific information. And the image-like input can be easily incorporated into degraded images or network layers to facilitate degradation-aware restoration.

\paragraph{Prompt Initialization.}
The initialization of visual prompts can be approached in various ways. In our case, since we consider them as visible images with the additive property, we choose a simple initialization method using one \texttt{nn.Parameter()} layer. This allows us to treat the visual prompts as trainable parameters that can be optimized during the restoration process.

To ensure training stability and expedite convergence, we employ a lightweight pre-trained image restoration model, \ie, MPRNet~\cite{DBLP:conf/wacv/MehriAS21}. We utilize this model to optimize the visual prompts by training them alongside the corresponding task datasets. In this process, we freeze all the weights of the pre-trained model, except for the parametric prompts, which are allowed to be learnt during training. This approach enables efficient optimization of the prompts while leveraging the knowledge encoded in the pre-trained model.

\paragraph{Prompt as Control.}
% thc revise here.
As previously discussed, we leverage the benefits of image-like prompts, allowing us to easily incorporate them into the original degraded images. 
By simply adding the prompts to the degraded images, we can effectively control
the restoration process.
Specifically, the control ability of degradation-aware prompts is mainly reflected in three aspects:
(1) the degradation-aware prompt guides the restoration for its corresponding degradation;
(2) the irrelevant prompts have no impact on the input without corresponding degradation;
(3) different prompts can be combined for complicated degradation, \eg, the input contains several degradation.
The capabilities in the above three aspects enable the proposed degradation-aware visual prompt to obtain the desired outputs.

In Fig. \ref{fig:S1_independent}, we illustrate some samples of combined images with various types of degradation. With the utilization of \name{}, we can observe that the desired images are successfully generated based on the provided visual prompts. \name{} demonstrates its capability to handle degraded images with prompts that may seem irrelevant or combined. Furthermore, we also try combinations of two visual prompts, the results with different weights of combination in Fig. \ref{fig:S3_combine} confirm the control ability of \name{}. For additional results, please refer to Sec.~\ref{control}.

\subsection{\name}
\paragraph{Architecture.}
The overall architecture of \name{} is illustrated in Fig.~\ref{fig:main}. Given the degraded image, we adopt a single task-specific visual prompt and add it to the input image and form the prompted image. We leverage a vanilla Vision Transformer (ViT)~\cite{DBLP:conf/iclr/DosovitskiyB0WZ21} pre-trained by MAE~\cite{DBLP:conf/cvpr/HeCXLDG22} as the encoder and a 2-conv pixel decoder to reconstruct the output RGB image. The intentionally simple design of the architecture is not intended to achieve top performance on various benchmarks, but rather to serve as a universal baseline for exploring the potential of degradation-aware visual prompts.

\paragraph{Training Loss.}
As the training process can be supervised by regression losses, we prefer simple losses such as $\ell_1$, $\ell_2$, and Smooth-$\ell_1$. We use Smooth-$\ell_1$ loss as $\mathcal{L}_{\text{pixel}}$ without sophisticated designs and it is effective enough for \name{}. For the performance of different training losses, we evaluate in ablation experiments in Sec.~\ref{training_loss}.

\subsection{Adaption via Prompt Tuning}
Previous works tend to fine-tune the whole models to adapt to new tasks or new datasets.
However, directly fine-tuning a new task is prone to lose the model's ability on the original task, and fine-tuning the whole model brings significant training costs.
In contrast, we freeze the weights of \name{} which is pre-trained on various tasks, and randomly initialize a new visual prompt for the new dataset or new task, as shown in Fig.~\ref{fig:main} (b).
During prompt tuning, we only update the learnable parameters of visual prompts by gradient descent, which is efficient without updating the entire model or long-schedule training. For prompt tuning, we can simply replicate the training settings used from scratch, ensuring that the prompts are optimized effectively without the need for extensive modifications or additional training procedures.

\section{Experiments}
\subsection{Datasets and Pre-processing}

Image restoration tasks are evaluated on several popular benchmarks, including SIDD~\cite{DBLP:conf/cvpr/AbdelhamedLB18} for image denoising, LoL~\cite{DBLP:conf/bmvc/WeiWY018} for low-light image enhancement, the merged deraining dataset~\cite{DBLP:journals/pami/ZamirAKHKYS23} for deraining, and the merged deblurring dataset~\cite{DBLP:conf/cvpr/ZamirA0HK0021}. A brief description of each dataset is provided below, and their details are summarized in Tab.~\ref{tab:datasets}.

\begin{wraptable}[20]{r}{0.4\linewidth}
\centering
\small
\vspace{-10pt}
\scalebox{0.9}{\begin{tabular}{l|l|l}
\toprule
Dataset & Train set & Test set \\
\midrule
\multicolumn{3}{l}{\textbf{Denoising}} \\
\hspace{5mm}SIDD \cite{abdelhamed2018high} & 320  & 1280  \\
\midrule
\multicolumn{3}{l}{\textbf{Low-light enhancement}} \\
\hspace{5mm}LOL \cite{wei2018deep} & 485 & 15 \\
\midrule
\multicolumn{3}{l}{\textbf{Merged Deraining}  \cite{DBLP:journals/pami/ZamirAKHKYS23}} \\
\hspace{5mm}Rain800 \ &  700 & 100 \\
\hspace{5mm}Rain1800  &  1800 & {-}  \\
\hspace{5mm}Rain14000  & 11200 & 2800  \\
\hspace{5mm}Rain1200  & {-} &  1200  \\
\hspace{5mm}Rain12  &  12 & {-} \\
\hspace{5mm}Rain100H  & {-} & 100  \\
\hspace{5mm}Rain100L & {-} & 100  \\
\midrule
\multicolumn{3}{l}{\textbf{Merged Deblurring} \cite{DBLP:conf/cvpr/ZamirA0HK0021}} \\
\hspace{5mm}GoPro \cite{nah2017deep} & 2103 & 1111  \\
\hspace{5mm}HIDE \cite{kim2019hide} & {-} & 2025 \\
\hspace{5mm}RealBlur-R \cite{rim2020real} & {-} & 980  \\
\hspace{5mm}RealBlur-J \cite{rim2020real} & {-} & 980 \\
\bottomrule
\end{tabular}}
\caption{Summary of the datasets used for \name{}.}
\label{tab:datasets}
\end{wraptable}

The Smartphone Image Denoising Dataset (SIDD)~\cite{abdelhamed2018high} is a large-scale dataset designed for image denoising. It contains both noisy and clean images captured by various smartphone cameras under different conditions, covering a diverse range of scenes and noise levels.

The Low-Light enhancement dataset (LOL)~\cite{wei2018deep} is designed for the task of low-light image enhancement. It contains 500 image pairs, with each pair consisting of a low-light image and its corresponding well-exposed ground truth. The images in the LOL dataset cover various indoor and outdoor scenes, with different levels of lighting and noise. 

The merged deraining dataset is obtained from the work of Li \etal~\cite{li2020learning}, which combines three popular deraining datasets: Rain100H~\cite{yang2017deep}, Rain100L~\cite{yang2017deep}, and Rain800~\cite{zhang2018density}. It is a comprehensive benchmark for evaluating single-image deraining algorithms. The merged dataset covers a diverse range of rain densities and streak orientations, providing a robust evaluation platform for assessing the performance of deraining algorithms.

The merged deblurring dataset is sourced from the work of Zhang \etal~\cite{zhang2021multistage}, which combines the GoPro dataset~\cite{nah2017deep} and three additional deblurring datasets: HIDE~\cite{kim2019hide}, RealBlur-R, and RealBlur-J~\cite{rim2020real}. This dataset contains various types and levels of blur caused by camera shake and object motion, making it a comprehensive and challenging benchmark for assessing the performance of deblurring algorithms.

\subsection{Training Details}
To learn prompts, we select a small image restoration model for quick fine-tuning. We use MPRNet~\cite{DBLP:conf/wacv/MehriAS21} and utilize the pre-trained models offered by the authors. We add one layer (\texttt{nn.Parameter()}) to the inputs, and freeze all other layers to ensure the prompts can be learnt during the fine-tuning stage. We reduce the learning rate to 1e-4 and remain other settings the same as training \name{}. 

To train the universal model, we choose a variant of ViT-Large according to~\cite{painter}. We employ the AdamW optimizer~\cite{AdamW} with a cosine learning rate scheduler, and train for 100 epochs. The training hyper-parameters are: the batch size as 160, base learning rate as 1e-3, weight decay as 0.05, $\beta_1$ = 0.9, $\beta_2$ = 0.999, drop path \cite{DBLP:conf/eccv/HuangSLSW16} ratio as 0.1, a warm-up for 2 epochs. We follow a light data augmentation strategy: random resize cropping with a scale range of [0.3, 1] and an aspect ratio range of [3/4, 4/3], with a random flipping. To make use of the pre-trained ViT-Large model, we resize the input image to $448\times448$. Considering that there are four different tasks, the sampling weight for each task is 0.3 (image deraining), 0.3 (low-light enhancement), 0.1 (image denoising), and 0.3 (image deblurring). Essential ablation studies are conducted to explore the effectiveness of some training strategies.

\subsection{Performance on Image Restorations Tasks}
With the corresponding task prompts, we compare our approach, namely \name{}, with recent best universal models and task-specific models on four representative image restoration vision tasks, shown in Tab. \ref{tab:compare}. Without task-specific design and only utilizing Smooth-$\ell_1$ loss for supervision, \name{} outperforms the universal models such as Painter~\cite{painter} and also gets closer results compared with the state-of-the-art task-specific model on several tasks. There is still much room for boosting our approach compared to other well-designed task-specific models. For example, our default training iteration number is 81k (50 epochs), while those task-specific models use a much larger number for training, \eg, 300k in MIRNet-v2 for image denoising, 150k for low-light enhancement, and 400k in MPRNet for all tasks. Achieving state-of-the-art performance on every task is not the goal of this paper, the unified model is expected to yield superior performance with more comprehensive training. 

\begin{table}
    \centering
    \small
    \setlength{\tabcolsep}{6pt}
    \renewcommand\arraystretch{1.2}
    \scalebox{0.9}{
    \begin{tabular}{l|cc|cc|cc|cc}
    \toprule
             \multirow{3}{*}{Method} & 
             \multicolumn{2}{c|}{Denoising} &
             \multicolumn{2}{c|}{Deraining} &
             \multicolumn{2}{c|}{Enhancement} &
             \multicolumn{2}{c}{Deblurring}\\
             & 
             \multicolumn{2}{c|}{SIDD} &
             \multicolumn{2}{c|}{5 datasets} &
             \multicolumn{2}{c|}{LoL}  &
             \multicolumn{2}{c}{4 datasets}\\
             & PSNR $\uparrow$ & SSIM $\uparrow$ & PSNR $\uparrow$ & SSIM $\uparrow$  & PSNR $\uparrow$ & SSIM $\uparrow$ & PSNR $\uparrow$ & SSIM $\uparrow$ \\

            \midrule
            \multicolumn{9}{c}{\small{{Task-specific models}}} \\
            \midrule
            {{Uformer \cite{DBLP:conf/cvpr/WangCBZLL22}}} &  {39.89} & {0.960} & {-} & {-} & {-} & {-} & {32.31} & {0.941} 
            \\
            {{MPRNet \cite{DBLP:conf/wacv/MehriAS21}}}  & {39.71} & {0.958} & {32.73} & {0.921} & {-} & {-} & {33.67} & {0.948}
            \\
            {{MIRNet-v2 \cite{DBLP:journals/pami/ZamirAKHKYS23}}} & {39.84} & {0.959} & {-} & {-} & \textbf{24.74} & {0.851}& {-} & {-} 
            \\

            {{Restormer \cite{DBLP:conf/cvpr/ZamirA0HK022}}} &
            \textbf{40.02} & {0.960} & \textbf{33.96} & \textbf{0.935} & {-} & {-}& {32.32} & {0.935}
            \\

            {{MAXIM \cite{DBLP:conf/cvpr/TuTZYMBL22}}} & {39.96} & {0.960} & {33.24} & {0.933} & {23.43} & {0.863}& \textbf{34.50} & \textbf{0.954}
            \\
            % multi-task models
            % AIRNet & RMBN dataset inconsistent
            
            % specialized models
            % KBNet & IPT lacks of too much results.
            \midrule
            \multicolumn{9}{c}{\small{{Universal models}}} \\
            \midrule
            Painter \cite{painter}&  38.88 & 0.954 & 29.49 & 0.868 & 22.40 & 0.872 & {-} & {-}  \\
            \name{}     & 39.28 & \textbf{0.967} & 30.67 & 0.891 & 22.73 & \textbf{0.877} & 28.03 & 0.897 \\
    \bottomrule
    \end{tabular}}
    \vspace{3pt}
    \caption{Comparison with the universal models, and the recent best task-specific models on four representative low-level image restoration tasks. The backbone of \name{} and Painter is ViT-Large.}
    % \vspace{-20pt}
    \label{tab:compare}
\end{table}

\subsection{Universal Models \textit{v.s.} Task-specific Models}
Tab.~\ref{tab:vit} compares the performance between universal models and task-specific models. For a fair comparison, we adopt the same architecture of \name{} and train a series of task-specific models along with a prompt-free universal model. All models are based on ViT-Large with MAE pre-trained weights and trained with the same setting (50 epochs).
In comparison to task-specific models, \name{} can achieve competitive or better performance, especially on low-light enhancement (LoL).

In comparison to the vanilla universal baseline, \ie, ViT-Large trained with the joint dataset, \name{} still obtains similar results on denoising and deraining.
However, \name{} achieves significantly better performance on the other two tasks, \ie, enhancement and deblurring, which shows the superiority of \name{} for learning with different tasks.
Note that we can leverage visual prompts in \name{} for specific restoration tasks and generate the desired outputs, which is infeasible for the vanilla universal model.

\begin{table}[ht] 
    \centering
    \small
    \setlength{\tabcolsep}{8pt}
    \renewcommand\arraystretch{1.2}
    \scalebox{0.9}{
    \begin{tabular}{l|cc|cc|cc|cc}
    \toprule
             \multirow{3}{*}{Method} & \multicolumn{2}{c|}{Denoising} &
             \multicolumn{2}{c|}{Deraining} &
             \multicolumn{2}{c|}{Enhancement} &
             \multicolumn{2}{c}{Deblurring}\\
             &  \multicolumn{2}{c|}{SIDD} &
             \multicolumn{2}{c|}{5 datasets} &
             \multicolumn{2}{c|}{LoL}  &
             \multicolumn{2}{c}{4 datasets}\\
             & PSNR $\uparrow$ & SSIM $\uparrow$ & PSNR $\uparrow$ & SSIM $\uparrow$  & PSNR $\uparrow$ & SSIM $\uparrow$ & PSNR $\uparrow$ & SSIM $\uparrow$ \\

            \midrule
            \multicolumn{9}{c}{\small{{Task-specific models}}} \\
            \midrule
            \multirow{4}{*}{ViT-Large}
            %{{ViT-L for denoising}} 
            & \textbf{39.74} & 0.969 & {-} & {-} & {-} & {-}& {-} & {-} 
            \\
            %{{ViT-L for deraining}} 
            &  {-} & {-} & {29.95} & {0.879} & {-} & {-} & {-} & {-} 
            \\
            %{{ViT-L for enhancement}}  
            & {-} & {-} & {-} & {-} & {18.91} & {0.741} & {-} & {-}
            \\
            %{{ViT-L for deblurring}} 
            & {-} & {-} & {-} & {-} & {-} & {-}& {27.51} & {0.882}
            \\
            \midrule
            \multicolumn{9}{c}{\small{{Universal models}}} \\
            \midrule
            ViT-Large &  39.28 & \textbf{0.967} & \textbf{30.75} & \textbf{0.893} & 21.69 & 0.850 & 20.57 & 0.680
            \\
            \name{}     & 39.28 & \textbf{0.967} & 30.67 & 0.891 & \textbf{22.73} & \textbf{0.877} & \textbf{28.03} & \textbf{0.897} \\
    \bottomrule
    \end{tabular}
    }
    \vspace{4pt}
    \caption{Comparison between \name{} and the vanilla task-specific models based on ViT-Large.}
    \label{tab:vit}
\end{table}

\subsection{Control Ability}
\label{control}
Compared to the vanilla universal model, one core advantage of \name{} is to control desirable outputs with given prompts. To reveal the control ability of \name{}, we adopt the following settings to evaluate \name{}. \emph{Setting1}: for images with single degradation, we apply single task-aware prompts; \emph{Setting2}: for images containing single degradation, we offer single task-irrelevant prompts; \emph{Setting3}: for images containing several complex degradation, we linearly combine prompts for different tasks for combined task-relevant prompts.

\begin{figure}
  \centering
  \includegraphics[width=1.0\linewidth]{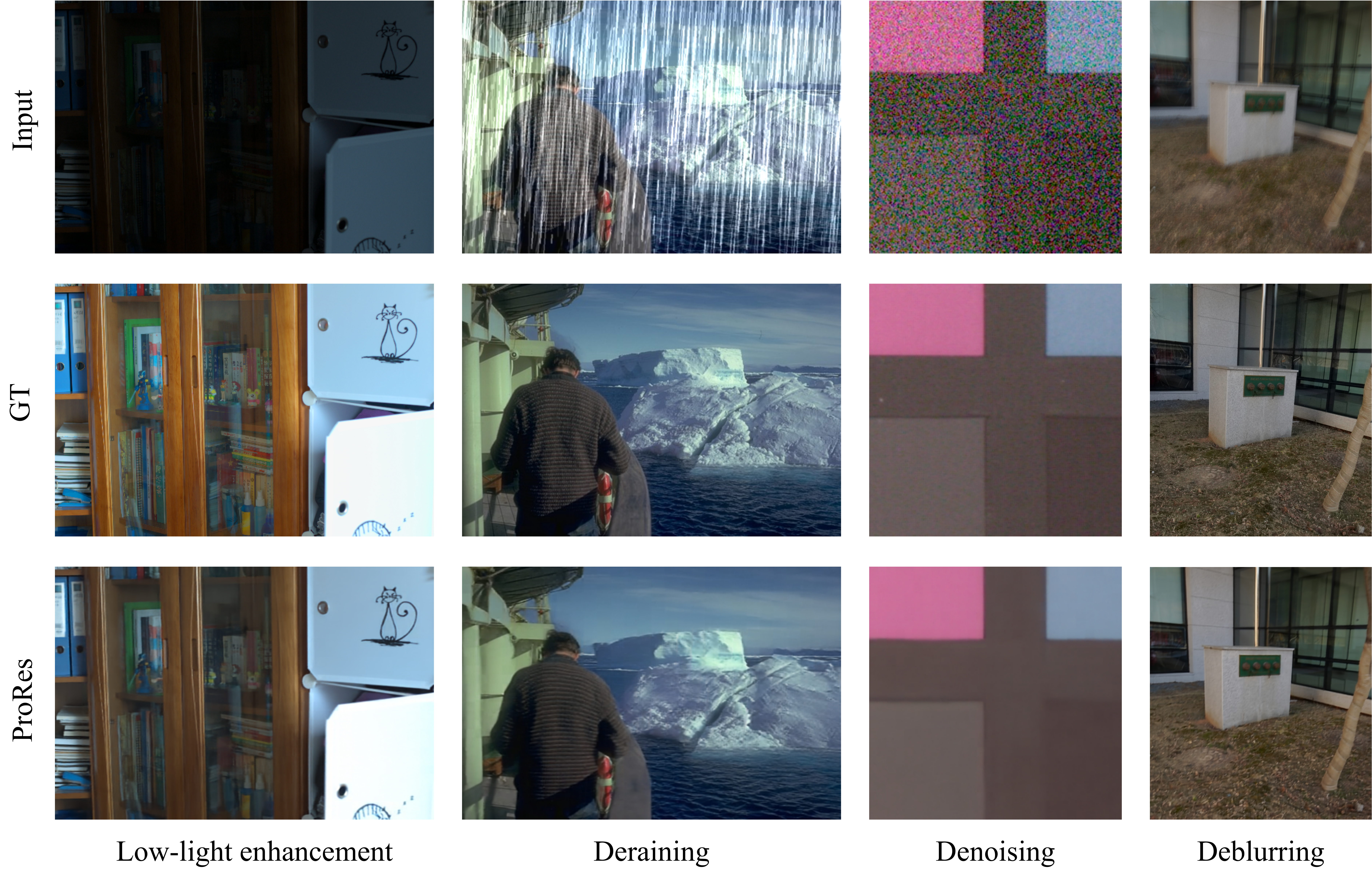}
  \vspace{-10pt}
  \caption{Visualization results processed from images of different corruptions. Compared with the original inputs, the outputs are consistent with the given visual prompts.}
  \label{fig:S1_independent}
\end{figure}

\paragraph{Independent Control.}
For \emph{Setting1}, we directly feed the original degraded images into our unified model with relevant prompts. As is shown in Fig. \ref{fig:S1_independent}, for those degraded images, we can easily control the outputs of our \name{} model by providing specific single prompts. It reflects the \name{} model's inherent ability for independent control.

\begin{figure}
  \centering
  \includegraphics[width=1.0\linewidth]{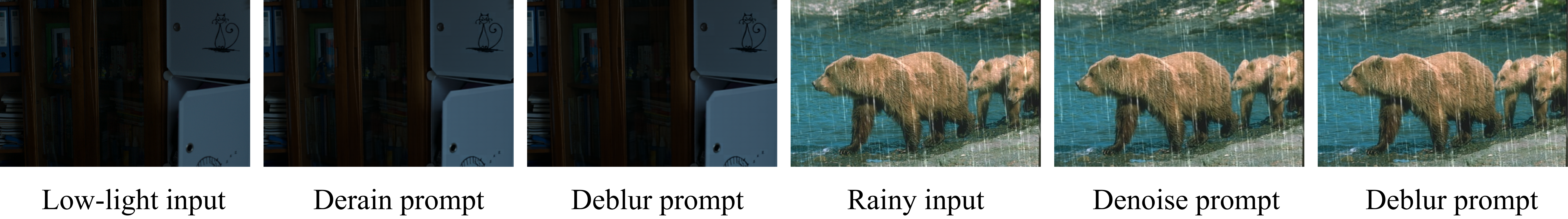}
  \vspace{-8pt}
  \caption{Visualization results processed by different prompts. Compared with the original inputs, the outputs remain unchanged with irrelevant visual prompts.}
  \label{fig:S2_irrelevant}
\end{figure}

\paragraph{Sensitive to Irrelevant Task-specific Prompts.}
In \emph{Setting2}, we directly feed the original degraded images into our unified model with random irrelevant prompts. As is shown in Fig. \ref{fig:S2_irrelevant}, the rainy images can still keep the same with the denoising prompt. It reflects that the \name{} model is sensitive to prompts from irrelevant tasks for each input image.

\paragraph{Tackle Complicated Corruptions.}
Regarding \emph{Setting3}, we explore the fusion of prompts and the ability to tackle complicated corruptions. We generate a subset from the Rain100L dataset, which is originally a deraining dataset. To ensure the subset contains several degradation, we follow the procedures of generating a synthetic low-light environment and making it a low-light rainy dataset. Considering that the subset does not exist paired ground truth for single degradation, hence we solely provide the qualitative evaluation. We combine those prompts by weights, and the aim is to jointly deal with those complicated corruptions. 

In Fig. \ref{fig:S3_combine}, we illustrate the results of different weights. We can find that the outputs can be controlled by adjusting the weight. Increasing the weight of low-light enhancement prompts will result in brighter output while increasing the weight of the deraining prompts will enhance the deraining effect.
By combining multiple prompts, \name{} can enhance the restoration performance and handle the challenging and diverse forms of corruption present in the images.

\subsection{Adaptation via Prompt Tuning}
Here, we evaluate the generalization ability and transferring ability of \name{} on new datasets or new tasks.
Specifically, we adopt the FiveK dataset~\cite{DBLP:conf/cvpr/BychkovskyPCD11} for low-light enhancement and the RESIDE-6K dataset~\cite{DBLP:conf/cvpr/AncutiAT20} for image dehazing, which is a new task for \name{}.
To adapt \name{} on new tasks or new datasets, we freeze all parameters of \name{} but the learnable visual prompts.
Specifically, we randomly initialize these two visual prompts and individually train on the two datasets for 50 epochs with all parameters of \name{} frozen, as shown in Fig.~\ref{fig:main} (b).

\begin{figure}
  \centering
  \includegraphics[width=1.0\linewidth]{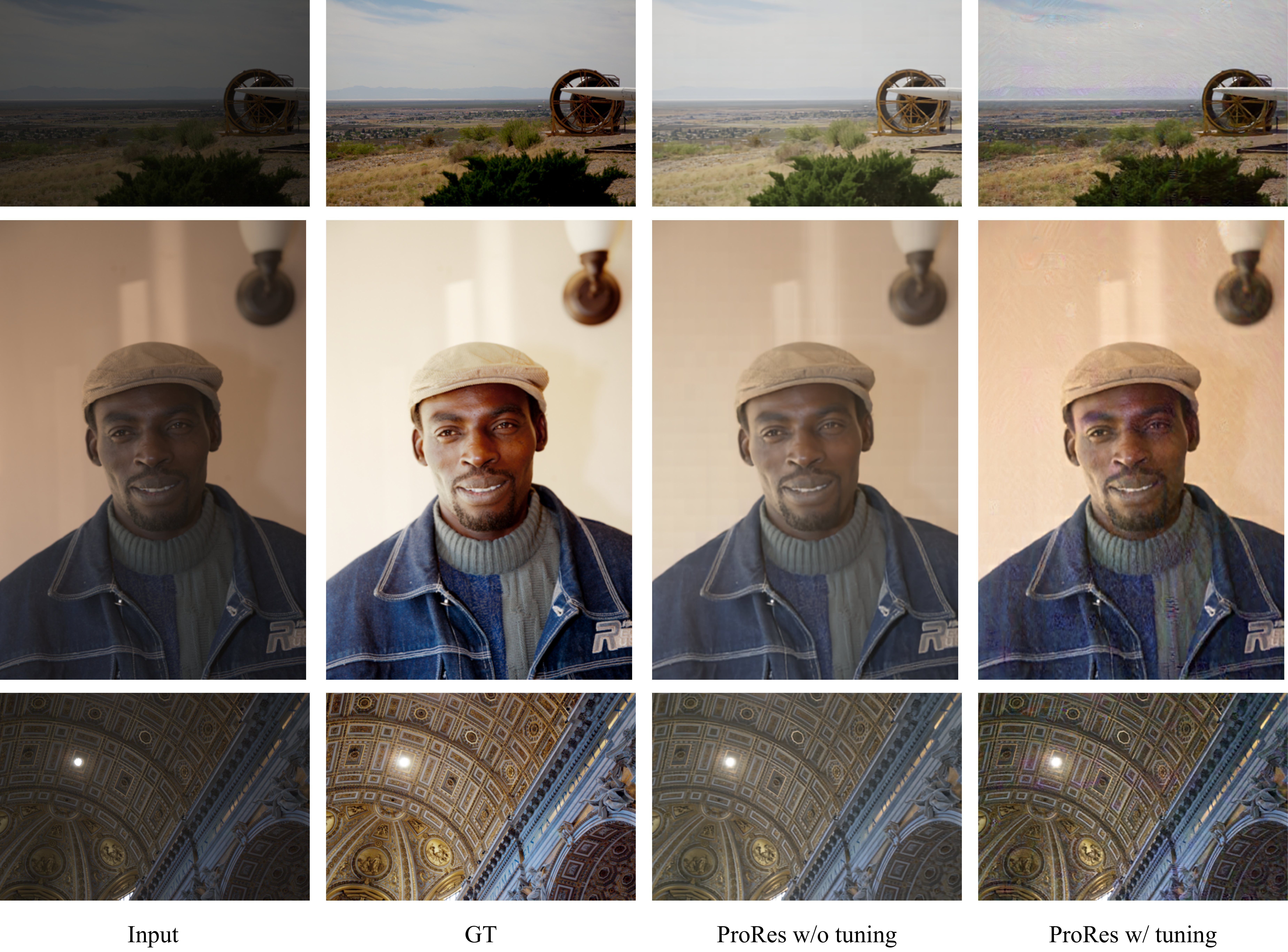}
  \vspace{-15pt}
  \caption{Visualization results of \name{} on the FiveK dataset. We adopt two settings, \ie, direct inference and prompt tuning, to evaluate \name{} on the FiveK dataset (a new dataset for low-light enhancement).}
  \label{fig:tuning_fivek}
\end{figure}

\begin{figure}
  \centering
  \includegraphics[width=1.0\linewidth]{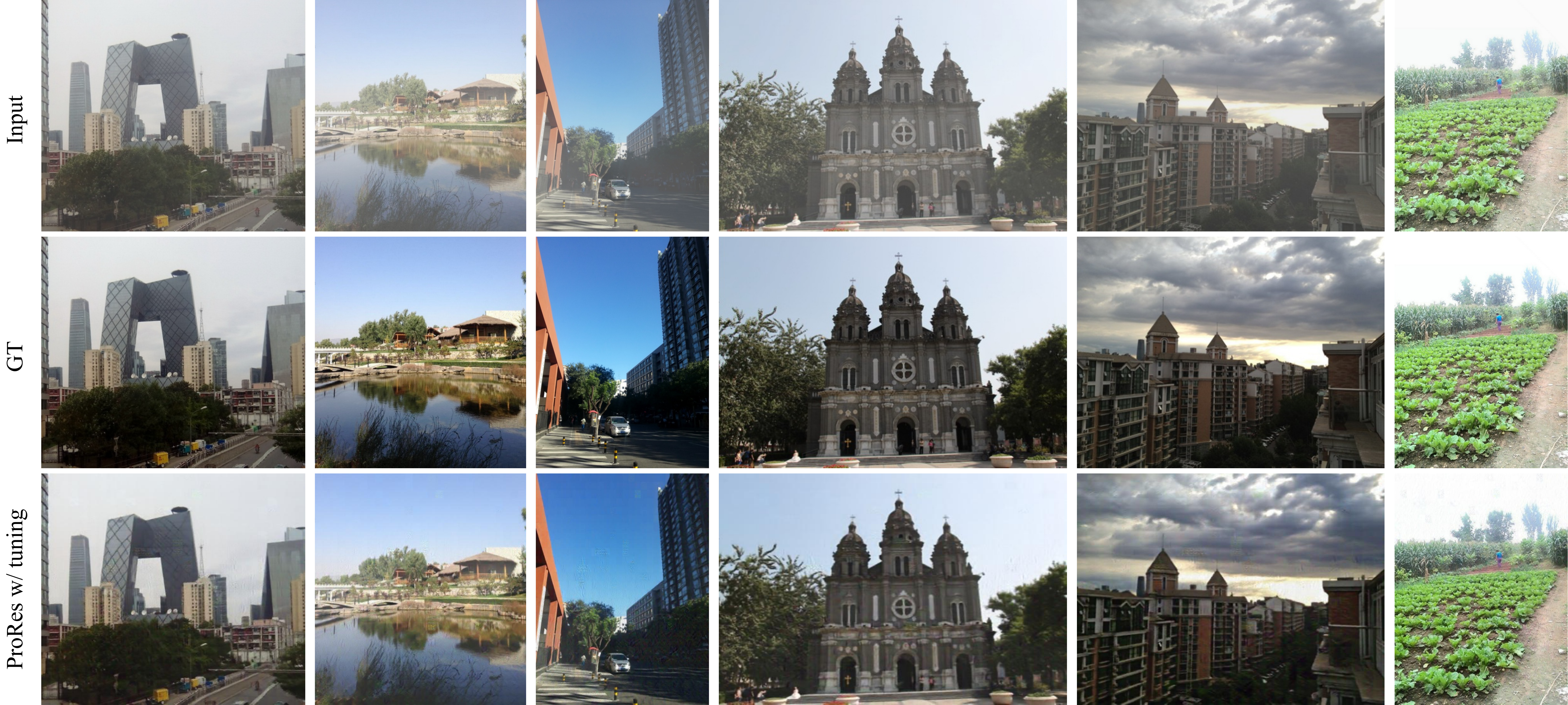}
  \vspace{-10pt}
  \caption{Visualization results of \name{} on the RESIDE-6K dataset via prompt tuning for image dehazing (a new task).}
  \label{fig:tuning_reside}
\end{figure}

In addition, we compare the performance of directly using \name{} and \name{} with prompt tuning.
As shown in Tab.~\ref{tab:prompt_tuning}, directly applying \name{} can achieve good performance on FiveK dataset, showing the generalization ability of \name{} on unseen samples.
Further, using prompt tuning brings remarkable improvements, which can demonstrate the effectiveness of prompt tuning and the transferring ability for new datasets (FiveK for low-light enhancement) or tasks (RESIDE-6K for dehazing).

\begin{table}
    \centering
    \small
    \setlength{\tabcolsep}{8pt}
    \renewcommand\arraystretch{1.2}
    \scalebox{0.9}{
    \begin{tabular}{l|cc|cc}
    \toprule
             \multirow{3}{*}{Method} & \multicolumn{2}{c|}{Enhancement} &
             \multicolumn{2}{c}{Dehazing} \\
             & \multicolumn{2}{c|}{FiveK} &
             \multicolumn{2}{c}{RESIDE-6K} \\
            & PSNR $\uparrow$ & SSIM $\uparrow$ & PSNR $\uparrow$ & SSIM $\uparrow$ \\
            \midrule

    \name{} w/o Prompt Tuning& 18.94 & 0.815 & - & -\\
    \name{} w/ Prompt Tuning & 22.78 & 0.839 & 21.47 & 0.840 \\
    \bottomrule
    \end{tabular}
    }
    \vspace{4pt}
    \caption{Experimental results of \name{} with prompt tuning on the FiveK and RESIDE-6K datasets.}
    \label{tab:prompt_tuning}
\end{table}

\subsection{Training \name{} with Degradation-aware Prompts}

In Tab.~\ref{tab:prompt}, we study the different strategies of training \name{} with degradation-aware visual prompts, \ie, initialization of visual prompts and prompt update.
We initialize \name{} with MAE~\cite{DBLP:conf/cvpr/HeCXLDG22} pre-trained weights and train \name{} from scratch.
We adopt two initialization methods for visual prompts: (1) random initialization (using \texttt{torch.nn.init.normal\_(self.prompt, std=0.1)}) and (2) pre-trained weights from light-weight image restoration models (default).
During training, the visual prompts can be detached without parameter update or learnable for the parameter update.
As shown in Tab.~\ref{tab:prompt}, using random initialization is inferior to pre-trained prompts in several tasks, \eg, deraining and enhancement.
However, training \name{} with learnable prompts brings negative impacts, and even leads to collapse in enhancement, which can be attributed to the limited training samples in enhancement.
Tab.~\ref{tab:prompt} can demonstrate that using detached pre-trained prompts is more stable and performs better.

\begin{table}[h]
    \centering
    \small
    \setlength{\tabcolsep}{6pt}
    \renewcommand\arraystretch{1.2}
    \scalebox{0.9}{
    \begin{tabular}{ll|cc|cc|cc|cc}
    \toprule
             \multicolumn{2}{c|}{\multirow{2}{*}{Prompt}} & \multicolumn{2}{c|}{Denoising} &
             \multicolumn{2}{c|}{Deraining} &
             \multicolumn{2}{c|}{Enhancement} &
             \multicolumn{2}{c}{Deblurring} \\
             & & \multicolumn{2}{c|}{SIDD} &
             \multicolumn{2}{c|}{5 datasets} &
             \multicolumn{2}{c|}{LoL}  &
             \multicolumn{2}{c}{4 datasets}\\
            Initialization & Learnable & PSNR $\uparrow$ & SSIM $\uparrow$ & PSNR $\uparrow$ & SSIM $\uparrow$  & PSNR $\uparrow$ & SSIM $\uparrow$ & PSNR $\uparrow$ & SSIM $\uparrow$ \\
            \midrule
            Random & Learnable & 39.24 & 0.966 & 29.98 & 0.881 & 10.60 & 0.417 & 26.19 & 0.844
            \\
            Random & Detached & 39.14 & 0.966 & 29.98 & 0.877 & 22.02 & 0.819 & \textbf{28.10} & \textbf{0.898}
            \\
            Pre-trained & Learnable & 39.26 & \textbf{0.967} & 30.20 & 0.884 & 22.47 & 0.876 & 27.83 & 0.891
            \\
            Pre-trained & Detached & \textbf{39.28} & \textbf{0.967} & \textbf{30.67} & \textbf{0.891} & \textbf{22.73} & \textbf{0.877} & 28.03 & 0.897 
            \\
    \bottomrule
    \end{tabular}
    }
    \vspace{4pt}
    \caption{Comparison of using different training strategies for \name{} with degradation-aware visual prompts.}
    \label{tab:prompt}
\end{table}

\subsection{Training Loss}
\label{training_loss}
Although \name{} is utilized for tackling low-level vision tasks, we opt to use straightforward regression losses rather than complex combinations of losses (Charbonnier loss \cite{DBLP:conf/icip/CharbonnierBAB94} or Perceptual loss \cite{DBLP:conf/eccv/JohnsonAF16}). Therefore, we only compare $\ell_1$, $\ell_2$, and Smooth-$\ell_1$ losses to provide clear guidance. As shown in Table \ref{tab:loss}, we find that Smooth-$\ell_1$ slightly outperforms $\ell_1$, and $\ell_2$ experiences a performance decline in most image restoration tasks. Hence, we use Smooth-$\ell_1$ as the training loss.

\begin{table*}[h]
    \centering
    \small
    \setlength{\tabcolsep}{6pt}
    \renewcommand\arraystretch{1.2}
    \scalebox{0.9}{
    \begin{tabular}{c|cc|cc|cc|cc}
             \toprule
             \multirow{3}{*}{Loss}& \multicolumn{2}{c|}{Denoising} &
             \multicolumn{2}{c|}{Deraining} &
             \multicolumn{2}{c|}{Enhancement} &
             \multicolumn{2}{c}{Deblurring}\\
             & \multicolumn{2}{c|}{SIDD} &
             \multicolumn{2}{c|}{5 datasets} &
             \multicolumn{2}{c|}{LoL} &
             \multicolumn{2}{c}{4 datasets}\\
              &  PSNR $\uparrow$ & SSIM $\uparrow$ & PSNR $\uparrow$ & SSIM $\uparrow$  & PSNR $\uparrow$ & SSIM $\uparrow$ & PSNR $\uparrow$ & SSIM $\uparrow$ \\
            \midrule
            $\ell_2$  & 38.93 & 0.956 & \textbf{30.69} & \textbf{0.893} & 22.48 & 0.873 & 27.91 & 0.896\\
            $\ell_1$ & 39.27 & \textbf{0.967} & 30.63 & 0.890 & 22.60 & 0.875 & \textbf{28.03} & \textbf{0.898}\\
            Smooth-$\ell_1$ & \textbf{39.28} & \textbf{0.967} & 30.67 & 0.891 & \textbf{22.73} & \textbf{0.877} & \textbf{28.03} & 0.897 \\
            \bottomrule
        \end{tabular}
        }
        \caption{Ablation on different losses for \name{}.}
        \label{tab:loss}
\end{table*}

\subsection{Visualization of Degradation-aware Visual Prompts}
\begin{figure}
  \centering
  % \vspace{-15pt}
  \includegraphics[width=\linewidth]{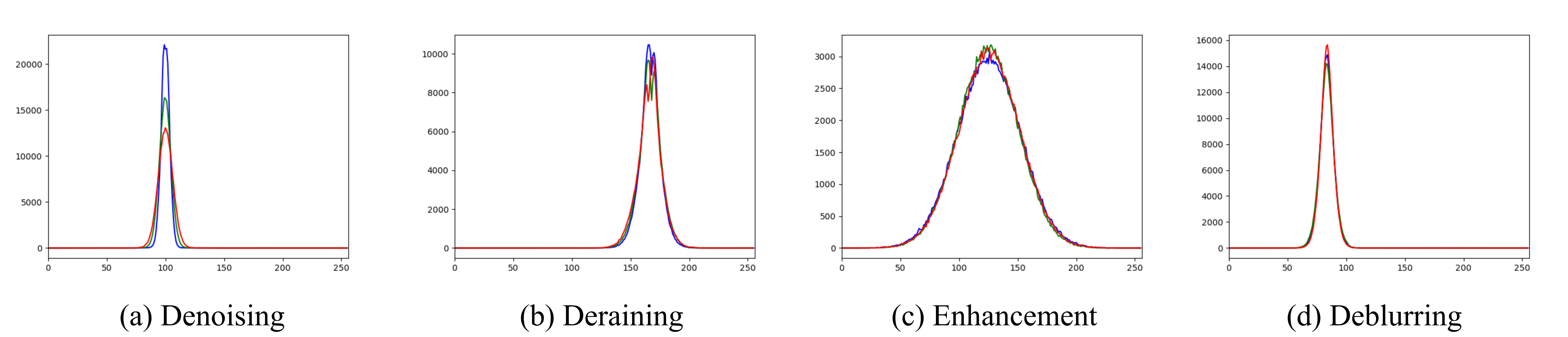}
  \vspace{-10pt}
  \caption{Histogram distribution of degradation-aware visual prompts. The red, blue, and green lines denote the 3 channels, \ie, RGB channels in input. It's clear to see that different visual  prompts have different distributions.}
  \label{fig:vis_prompt}
\end{figure}

Here, we reveal discriminative visual prompts by visualizing the histogram distribution of each visual prompt. Fig.~\ref{fig:vis_prompt} illustrates the distinct distributions of visual prompts for different degradation types. These notable differences in distribution highlight the control ability of visual prompts and the potential for weighted combinations of prompts in hybrid restoration tasks. By leveraging these distinct distributions, we can effectively manipulate the restoration process and achieve desired outcomes based on the specific degradation types and requirements.

\subsection{Integrate Prompts in Different Layers}
To better reveal the compatibility of prompts in different layers of the model, we design several settings which are trained from scratch. Besides directly adding visual prompts to input, we try to integrate them into layers in \name{} network. We integrate visual prompts in all layers, equal spacing, the first 5 layers, and the last 5 layers.

\begin{table*}[h]
    \centering
    \small
    \setlength{\tabcolsep}{6pt}
    \renewcommand\arraystretch{1.2}
    \scalebox{0.9}{
    \begin{tabular}{c|cc|cc|cc|cc}
             \toprule
             \multirow{3}{*}{Layers} & \multicolumn{2}{c|}{Denoising} &
             \multicolumn{2}{c|}{Deraining} &
             \multicolumn{2}{c|}{Enhancement} &
             \multicolumn{2}{c}{Deblurring}\\
             & \multicolumn{2}{c|}{SIDD} &
             \multicolumn{2}{c|}{5 datasets} &
             \multicolumn{2}{c|}{LoL} &
             \multicolumn{2}{c}{4 datasets}\\
              &  PSNR $\uparrow$ & SSIM $\uparrow$ & PSNR $\uparrow$ & SSIM $\uparrow$  & PSNR $\uparrow$ & SSIM $\uparrow$ & PSNR $\uparrow$ & SSIM $\uparrow$ \\
            \midrule
            Add to input & \textbf{39.28} & \textbf{0.967} & 30.67 & 0.891 & \textbf{22.73} & 0.877 & \textbf{28.03} & \textbf{0.897} \\
            \midrule
            All   & 39.16 & 0.966 & 30.31 & 0.884 & 22.67 & 0.875 & 26.83 & 0.867 \\
            0, 5, 11, 17, 23 & 39.25 & \textbf{0.967} & 30.57 & 0.888 & 22.47 & 0.873 & 27.95 & 0.896 \\
            0, 1, 2, 3, 4 & 39.26 & \textbf{0.967} & 30.49 & 0.888 & 23.15 & 0.882 & 27.99 & 0.896 \\
            19, 20, 21, 22, 23 & 39.27 & \textbf{0.967} & \textbf{30.71} & \textbf{0.892} & 22.70 & \textbf{0.884} & 20.57 & 0.680 \\
            \bottomrule
        \end{tabular}
        }
        \caption{Quantitative results of different prompt integration strategies.}
        \label{tab:combine}
\end{table*} 

In Tab. \ref{tab:combine}, we can find that \name{} can be effective in almost all settings, except the last setting, \ie, the last 5 layers. With only the last five layers integration, the performance on deblurring drops dramatically. This phenomenon is also revealed in other settings compared with only adding visual prompts to input. Hence, we uphold the idea that simplicity is the best and choose the simplest strategy, \ie, add to input.

\section{Conclusion}

In this paper, we propose a universal framework named \name{} for versatile image restoration, which leverages the presented degradation-aware visual prompts as task identifiers. \name{} adopts the simple transformer architecture without task-specific designs and obtains competitive performance on various benchmarks for image restoration.
Extensive experiments demonstrate the control ability with combined prompts and transfer ability on new tasks through prompt tuning.
We believe \name{} is a significant step towards controllable and versatile image restoration and can motivate future research.

%%%%%%%%%%%%%%%%%%%%%%%%%%%%%%%%%%%%%%%%%%%%%%%%%%%%%%%%%%%%

\end{document}